\documentclass[letterpaper, 10 pt, conference]{ieeeconf}  

\IEEEoverridecommandlockouts                              

\usepackage{graphics}  
\usepackage{epsfig}    
\usepackage{amsmath}   
\usepackage{bm}
\usepackage{mathtools}
\usepackage{commath}
\usepackage{color}
\usepackage{soul}
\usepackage{amssymb}
\usepackage{comment}
\usepackage{xfrac}
\usepackage{nicefrac}

\usepackage{hyperref}
\usepackage{url}
\usepackage[noadjust]{cite}
\usepackage{xcolor}
\usepackage[percent, abs]{overpic}

\usepackage[font=small]{caption}
\usepackage[font=small]{subcaption}

\usepackage{algorithm}
\usepackage{algorithmic}

\usepackage[bottom]{footmisc}
\usepackage{gensymb}
\usepackage{graphicx}
\graphicspath{ {C:/Users/work/Desktop/PhD Work/latex work/figures/} }
\renewcommand{\baselinestretch}{0.97}

\usepackage{multirow, makecell, booktabs}

\title{\LARGE \bf
A Strawberry Harvesting Tool with Minimal Footprint
}

\author{Mohamed Sorour$^{1}$, Mohamed Heshmat$^{2}$, Khaled Elgeneidy$^{3}$ and Pål Johan From$^{4}$ 
\thanks{$^{1}$Insect Robotics Group - Institute for Perception, Action and Behaviour,
School of Informatics, University of Edinburgh. Informatics Forum, 10 Crichton St, EH8 9AB Edinburgh, United Kingdom.\newline%
{\tt\small msorour@ed.ac.uk}\newline \indent%
$^{2}$Research Office, Mohamed bin Zayed University of Artificial Intelligence, Abu Dhabi, United Arab Emirates.\newline \indent%
$^{3}$School of Engineering, Coventry University, Egypt branch, R7 New Administrative Capital, Cairo, Egypt.\newline \indent%
$^{4}$Robotics Group, Faculty of Science and Technology, Norwegian University of Life Sciences (NMBU), 1432 Ås, Norway.\newline \indent%
}}

\begin{document}

\maketitle
\thispagestyle{empty}
\pagestyle{empty}

\begin{abstract}
In this paper, a novel prototype for harvesting table-top grown strawberries is presented, that is minimalist in its footprint interacting with the fruit. In our methodology, a smooth trapper manipulates the stem into a precise groove location at which a distant laser beam is focused. The tool reaches temperatures as high as $188^{\degree}$ Celsius and as such killing germs and preventing the spread of local plant diseases. The burnt stem wound preserves water content and in turn the fruit shelf life. Cycle and cut times achieved are $5.56$ and $2.88$ seconds respectively in successful in-door harvesting demonstration. Extensive experiments are performed to optimize the laser spot diameter and lateral speed against the cutting time.


\end{abstract}


\section{Introduction}
Due to the unpredictable and seasonal characteristics of hand-picked crop collection \cite{Nolte2017,uk_gouv}, rising workforce requirements across competitive industries \cite{workforce_in_agriculture_world_bank,nfu_labour_availability_issues}, and the growing demographic challenge of an aging agricultural workforce \cite{Duckett2018}, the urgency for robotic harvesting solutions is widely recognized and validated. This persistent gap in automation contributes to heightened living expenses \cite{Cassey2018}, driven by inflated labor acquisition costs, alongside reduced agricultural output caused by crop losses from delayed harvesting. Conversely, existing robotic harvesting systems demonstrate moderate success rates averaging around $66\%$ \cite{Bac2014}, which further declines significantly in dense field conditions, a limitation frequently linked to inefficient, cumbersome harvesting tools \cite{Kootstra2021}.

\begin{figure}[t!]
    \centering
    \includegraphics[width=1.0\linewidth]{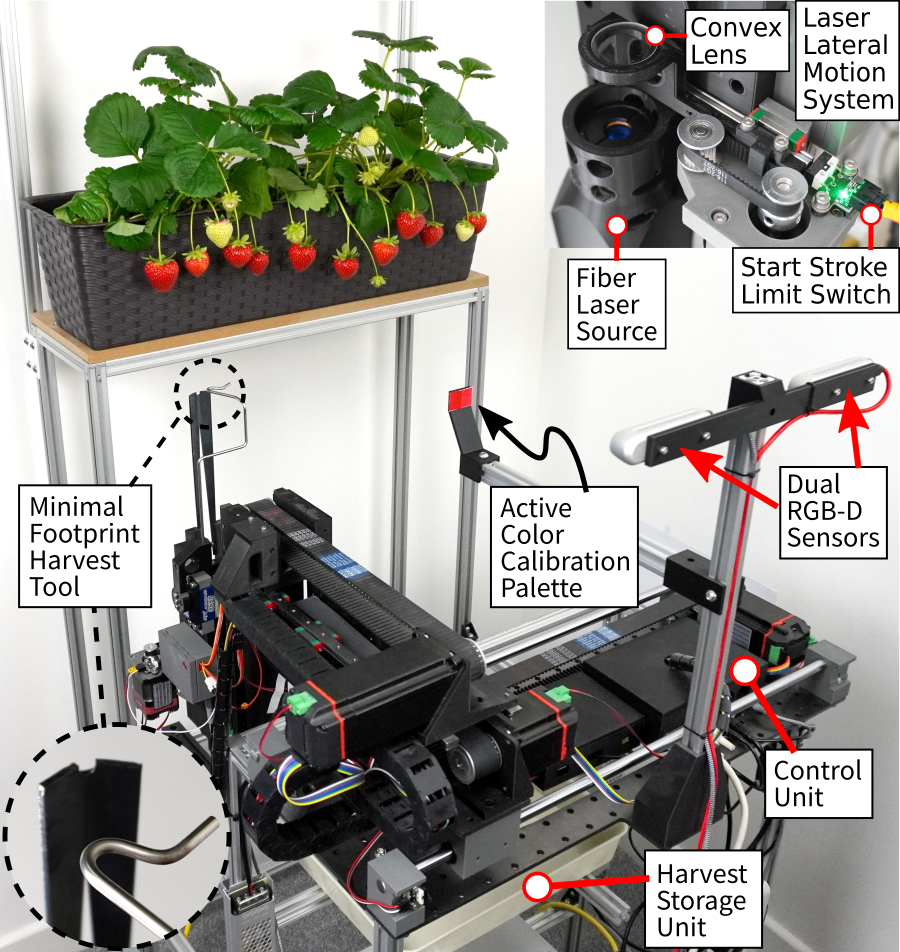}
    \captionsetup{aboveskip=10pt}
    \captionsetup{belowskip=-10pt}
    \caption{Belt driven Cartesian motion system for table-top strawberry harvesting.}
    \label{figure_setup_overview}
\end{figure}

Manual "\textit{human}" strawberry pickers are reported to have a cycle time as low as $1.2$s (seconds) per fruit, however this can only be maintained for no longer than $4$ hours \cite{Woo2020}, as such we believe a robotic harvesting rate of $6$s per fruit breaks even with human labor rate assuming a $20$ hour robot picker shift (to accommodate for battery charging and regular maintenance) in terms of productivity. Most of the developed tools in the literature of selective harvesting \cite{Soran_2024,Tituana_2024,Ezekyel_2024,Chang_2024,Li_2023,Xiong_2020,Andreas_2018, Zhou2022} share at least one of two features: (1) a sort of mechanical grasping system, and (2) a mechanical cutting tool. Both leading to a too bulky system for a delicate task to be done in often cluttered environment. Non-conventional cutting means on the other hand are rare in the literature, featuring an oscillating blade to cut sweet pepper in \cite{Lehnert2017}, and a thermal cutting device in \cite{vanHenten2002,Bachche2013} for cucumber and pepper respectively. Laser beam is used in \cite{Liu2008,Liu2011} to showcase the potential for cutting tomato peduncles, and in \cite{Heisel2002, Mathiassen2006, Coleman2021, Nadimi2021} for weed control. Despite the non-conventional cutting, bulky grasping is employed in the aforementioned works.

\begin{figure*}[t!]
    \centering
    \includegraphics[width=\textwidth]{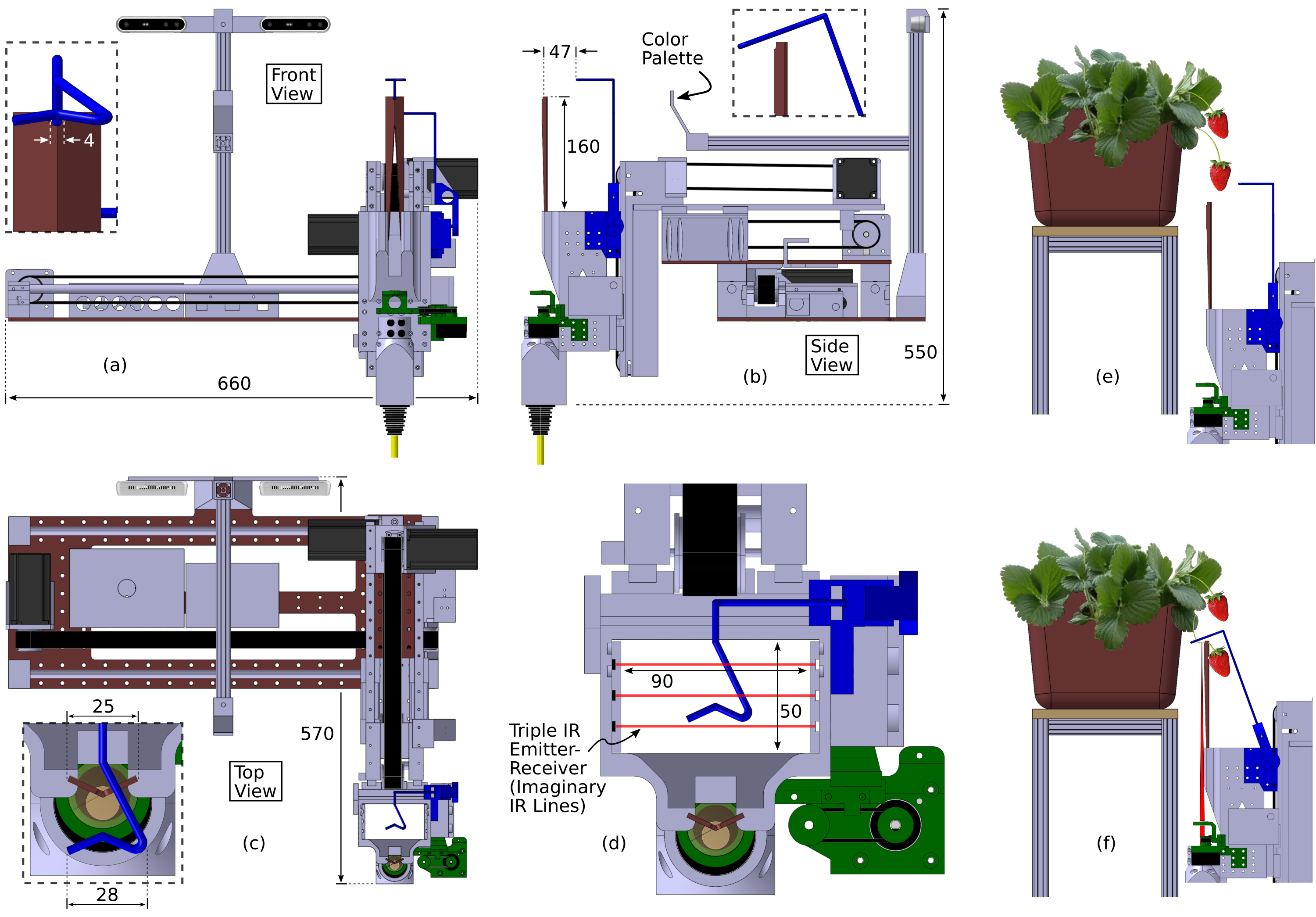}
    \captionsetup{aboveskip=10pt}
    \captionsetup{belowskip=-10pt}
    \caption{The developed harvesting tool setup in front (a), side (b), and top (c) views, as well as a respective close up views to the trapper tool in dotted rectangle. Enlarged tool top view in (d) showing the imaginary infrared lines. Strawberry harvesting schematic demonstration in (e) and (f). Selective dimensions shown in millimeter.}
    \label{figure_harvesting_tool_CAD}
\end{figure*}

In this work, the authors present a strawberry harvesting tool prototype designed for minimal interaction with the fruit and its environment, depicted in Fig. \ref{figure_setup_overview}. As in previous work \cite{Sorour2022, Sorour2022_IROS,Sorour2025}, we employ a focused laser beam for distant cutting, and use here a smooth stem trapping mechanism that minimizes friction during stem manipulation into the trap. The harvesting cycle and cutting times are $5.56$s and $2.88$s respectively. Laser cutting is as such slower than mechanical cutting tools, but it allows for a less bulky tool design, eliminates transmission of plant diseases due to the high temperatures involved, and requires far less maintenance. In addition, we present extensive experiments in this work to determine the impact of 1) laser spot motion dynamics and 2) laser spot diameter on the time of cut.
This paper is organised as follows, section II introduces the harvesting system and the operation logic. Experiments to optimize the laser spot motion and diameter are detailed in section III. The strawberry harvesting demonstration and associated algorithms are reported in section IV. Conclusions are finally delivered in section V. Videos to all experiments are hyperlinked in the sequel in the order of appearance for the reader's convenience.

\section{System Anatomy and Operation Logic}
The computer aided design (CAD) model of the harvesting system is shown in Fig. \ref{figure_harvesting_tool_CAD} in front Fig. \ref{figure_harvesting_tool_CAD}(a), side Fig. \ref{figure_harvesting_tool_CAD}(b) and top Fig. \ref{figure_harvesting_tool_CAD}(c) views, together with the respective close up views the trapper tool depicted in dotted rectangle. The Laser lateral motion, and stem trapping systems are colored in green and blue respectively throughout the figure. Laser spot motion is achieved by moving the convex lens to achieve high lateral speeds and power efficiency as opposed to moving the whole laser head as in \cite{Sorour2022}. This is equally effective given the required short stroke of $5mm$ for the laser spot. A protruding lever (more clearly visible in Fig. \ref{figure_setup_overview}) is designed to carry a printed color calibration palette or real fruit that lies in the vision cone of two mounted RGB-D cameras, such cone is omitted from the figure for a clear noise-free device depiction. A $3D$ Cartesian motion system is adopted here as opposed to the robot manipulation system in previous work by the authors in favor of shorter harvesting cycle time and cost effectiveness. In this work, the linear motion is belt driven and actuated using closed loop stepper servo system. Laser spot motion is achieved using a belt driven open loop stepping linear motion system actuating the convex lens. The initial reference position of the lens is marked using a limit switch, referencing is performed at the beginning of operation and after each single fruit harvesting to ensure accuracy is not compromised. The trapping system consists mainly of a polished chrome plated stainless steel rod with $3mm$ diameter, bent into $120\degree$ v-shape trapper. Total length of the bent rod is $300mm$ to allow for a distant trapping of the stem and by doing so protecting the fruit, solving a drawback in previous work. The trapper is actuated using mini servo motor and although controlled in open ($90\degree$ to the ground plane) or closed ($120\degree$) modes, it can be controlled to accommodate for various fruit sizes, however here we focus on strawberry harvesting and as such the aforementioned angular ranges are sufficient. The polished trapper rod forces the stem into a trapping groove, colored brown in Fig. \ref{figure_harvesting_tool_CAD} with minimal friction thanks to the smooth surface, the real world implementation of this system is shown in Fig. \ref{figure_setup_overview}. The trap groove is made of mild steel, bent to $120\degree$ in opposite configuration to that of the trapper rod. It has a groove of $4mm$ width in which the stem is trapped prior to laser cutting activation. The trapper rod and groove are $28mm$ and $25mm$ in width respectively, and as such have minimal interaction with the fruit as it targets only the stem. Such small width for the trapping grove is optimal for supporting the fruit during the laser cutting operation since the average diameter of the smallest strawberry cultivar is approximately $25mm$. The harvesting device has a working stroke of $48cm$ and is intended to be fitted to a mobile robot to cover a large area of the tabletop grown strawberry farm. A schematic CAD of the initial and final poses for single strawberry harvesting is shown in Fig. \ref{figure_harvesting_tool_CAD}(e) and Fig. \ref{figure_harvesting_tool_CAD}(f) respectively. Where the point cloud segmented fruit is approached from below at the localized $x$, $y$ axes coordinates, the tool is raised in $z$ axis to surpass the fruit height by $5cm$. The trapper is then closed, and the laser beam cutter is engaged as well as the laser spot motion system. When the stem is cut, the fruit falls into a storage bag, passing the triple infrared photo interrupter lines shown in Fig. \ref{figure_harvesting_tool_CAD}(d), at which point the laser beam disengages, the tool opens and moves on to the next fruit to cut.

\begin{figure}[t!]
    \centering
    \includegraphics[width=0.9\linewidth]{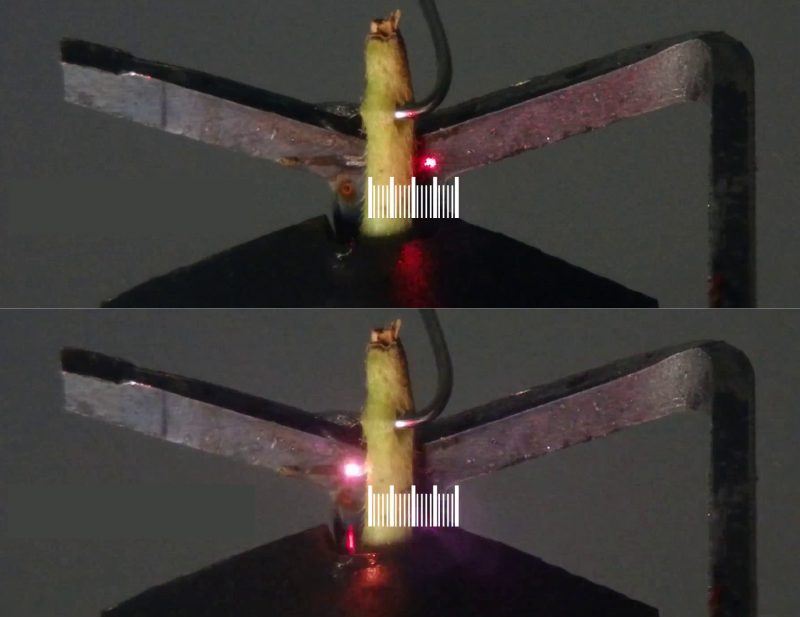}
    \captionsetup{aboveskip=10pt}
    \captionsetup{belowskip=-10pt}
    \caption{Experimental setup evaluating the effect of laser lateral velocity on the time of stem cut.}
    \label{figure_lateral_laser_speed_vs_stem_cut_speed}
\end{figure}

\section{Laser Beam Features and Time of Cut}
In this section we perform three experiments. the aim of the first is to determine the impact of laser beam lateral motion speed on the time of cut $t_c$. The second is to investigate the effect of laser spot diameter on the same metric, and the third aims to find the optimal laser spot diameter. Subsections below will detail the experimental setup of each and discuss the results listed in Tables \ref{table_lateral_laser_speed_vs_cut_velocity},\ref{table_lateral_spot_diameter_vs_cut_velocity}, and \ref{table_lateral_spot_diameter_optimization}.
\subsection{Laser Beam Motion Dynamics}
Firstly we perform an experiment to investigate the effect of cutting dynamics.
That is the effect of lateral laser beam movement velocity on the time of cut. Such motion dynamics is expected to impact the time of cut, since a too slow laser spot movement, for example: a single pass cut, would result in steady, but non uniform stem penetration. This will result in fruit/stem movement due to redistribution of the fruit weight on the uncut, load bearing part of the stem, and in turn longer cutting duration. Another point to consider, is that faster lateral motion might keep the stem water-content at a higher temperature, and as such faster to evaporate and cut. The general idea is that, much like pulse width modulation, a high speed lateral motion might generate a slower, but uniform stem cut.

\renewcommand{\arraystretch}{1.1} 
\begin{table}[b]
\captionsetup{aboveskip=-1pt,belowskip=-10pt}
\caption{Lateral laser velocity $v_l$ versus average stem cut velocity $\overline{v}_c$ at different laser spot diameters $\phi_{ls}$}
\label{table_lateral_laser_speed_vs_cut_velocity}
\begin{center}
\begin{tabular}{|c|c||c||c|c|}
\hline
$\phi_{ls}(mm)$ & $v_l(mm/s)$ & $\overline{\phi}_s(mm)$ & $\overline{t}_c(s)$ & $\overline{v}_c(mm/s)$ \\
\hline
\hline
\multirow{ 9}{*}{$0.1mm$} & $10$  & $2.27$  & $11.98$     & $0.19$ \\
& $20$  & $2.23$  & $11.42$     & $0.20$ \\
& $32$  & $2.15$  & $12.50$     & $0.17$ \\
& $40$  & $2.00$  & $12.58$     & $0.16$ \\
& $50$  & $2.16$  & $11.30$     & $0.19$ \\
& $59$  & $2.37$  & $13.62$     & $0.17$ \\
& $69$  & $2.39$  & $13.87$     & $0.17$ \\
& $78$  & $2.11$  & $12.64$     & $0.17$ \\
& $96$  & $2.25$  & $11.15$     & $0.20$ \\
\hline
\hline
$0.5mm$ & $96$  & $2.34$  & $07.33$     & $0.32$ \\

\hline
\end{tabular}
\end{center}
\end{table}

\begin{figure*}[t!]
    \centering
    \includegraphics[width=\textwidth]{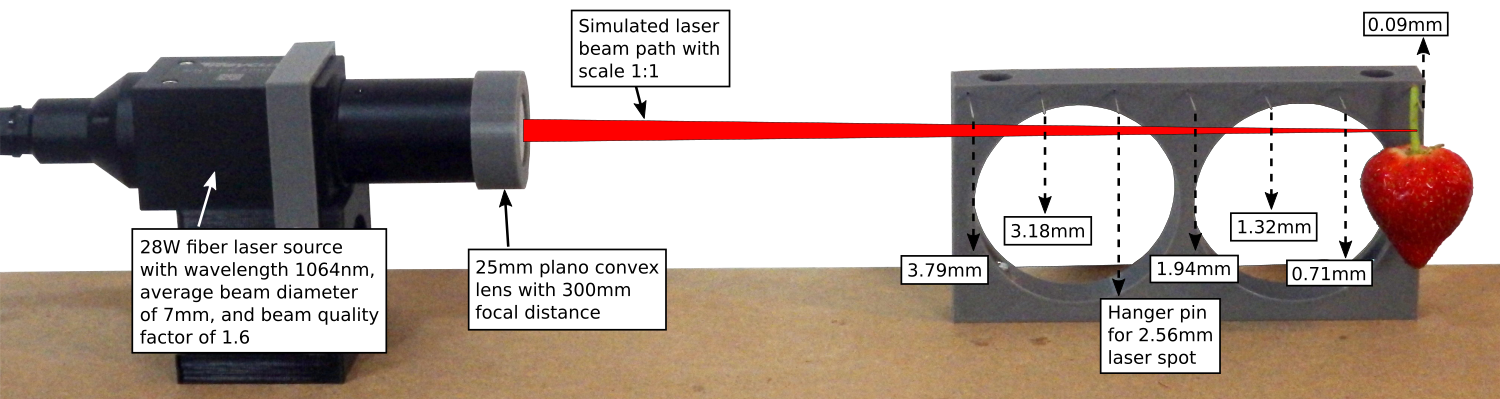}
    \captionsetup{aboveskip=10pt}
    \captionsetup{belowskip=-10pt}
    \caption{Setup for evaluating the impact of laser spot diameter on the stem piercing velocity, featuring $7$ hangers corresponding to differing spot diameters are tested for coarse tuning.}
    \label{figure_stem_piercing_setup}
\end{figure*}

The experimental setup used in this evaluation is shown in Fig. \ref{figure_lateral_laser_speed_vs_stem_cut_speed}, depicting the initial (upper) and final (lower) laser stroke positions (shown as red dot), as well as a calibrated ruler to measure the stem diameter per experiment. The stem is fixed to a hanger simulating a trapped stem between the tool trapper and the groove. A video summary of this experiment is submitted with this paper, also available online\footnote{\url{https://www.youtube.com/watch?v=ZBtjZAwG9Uo}}. In each experiment, the focused laser spot is moved laterally back and forth in a stroke of $4mm$ at a predetermined velocity. The strawberries used are selected and fitted midway the laser stroke, ensuring the laser spot completely passes each stem. Here, we tested $9$ different laser lateral velocity $v_l$ values. For each value, $10$ experiments are performed and the average stem diameter $\overline{\phi}_s$, time to cut $\overline{t}_c$, and cutting velocity $\overline{v}_c$ are recorded for a subtotal of $90$ trials in the first experiment. Initially the experiments were done at laser spot diameter $\phi_{ls}=0.1mm$, since a general conclusion from previous work \cite{Sorour2022_IROS} is that the smaller the sport diameter, the higher the energy density and in turn the stem penetration and cutting speed. A high quality plano-convex lens \cite{lens_edmund} is used with $300mm$ focal length and a $1\%$ focal length tolerance. The distance between the center plane of the convex lens and the trapper in Fig \ref{figure_lateral_laser_speed_vs_stem_cut_speed} is designed to result in a laser spot diameter of $0.1mm$ assuming the nominal value of focal length. The results were much slower than expected for aforementioned $90$ trials, so another pilot experiment was conducted at a single lateral velocity value for a laser spot diameter of $\phi_{ls}=0.5mm$, the results are listed in Table \ref{table_lateral_laser_speed_vs_cut_velocity} for a total of $100$ trials. In this table, we deduce that at $0.1mm$ laser spot diameter the average cutting velocity $\overline{v}_c$ didn't improve consistently with increasing the lateral laser speed. Such parameter is as such concluded to be of negligible contribution to the speed of cut. We do however expect that at lower velocities than those tested in the experiment would induce uneven stem cutting due to fruit movement and in turn lower cutting speeds. On the other hand, at $0.5mm$ laser spot diameter, we can observe a significant increase in the speed of cut. This contradicts a general prior conclusion that the smaller the laser spot diameter the better cutting speed achievable due to the higher energy density. The logical interpretation is that what matters in stem cutting is the area etching speed rather than the speed of stem piercing. The result as such suggests the existence of an optimized laser spot diameter to achieve faster overall cutting speed (area etching), finding such value motivates the following set of experiments.

\subsection{Optimal Laser Spot Diameter}

\renewcommand{\arraystretch}{1.1} 
\begin{table}[b]
\captionsetup{aboveskip=0pt,belowskip=-10pt}
\caption{Average stem piercing velocity $\overline{v}_p$ and the stem pierce constant $C_p$ at different laser spot diameters $\phi_{ls}$}
\label{table_lateral_spot_diameter_vs_cut_velocity}
\begin{center}
\begin{tabular}{|c|c||c||c|c|}
\hline
$\phi_{ls}(mm)$ & $\overline{\phi}_s(mm)$ & $\overline{t}_p(s)$ & $\overline{v}_p(mm/s)$ & $C_p(mm^2/s)$ \\
\hline
\hline
$3.79$ & $2.3$  & $32.0$  & $0.07$     & $0.26$ \\
$3.18$ & $2.3$  & $19.2$  & $0.12$     & $0.38$ \\
$2.56$ & $2.2$  & $8.92$  & $0.25$     & $0.64$ \\
$1.94$ & $2.3$  & $6.93$  & $0.34$     & $0.66$ \\
$1.32$ & $2.4$  & $2.61$  & $0.92$     & $1.21$ \\
$0.71$ & $2.3$  & $0.96$  & $2.42$     & $1.72$ \\
$0.09$ & $2.2$  & $0.71$  & $3.09$     & $0.28$ \\
\hline
\end{tabular}
\end{center}
\end{table}

Building upon the observations drawn from the first set of experiments, it is clear that despite a small spot diameter has higher energy density and as such faster stem piercing speed, it is not optimal in terms of stem cutting by gradual area etching. To determine the optimal laser spot diameter, two sets of experiments are performed for coarse and fine tuning respectively. The setup is shown in Fig. \ref{figure_stem_piercing_setup}, where the laser beam originating from the source to the left is focused through a high quality convex lens on route to a pre-hanged strawberry at certain distance from the lens corresponding to a laser spot diameter as shown in the respective figure. The focused laser beam is aimed at the strawberry stem at its midst, corresponding to the longest travel distance through the stem, until it penetrates and emerges from the other side. A to-scale focused laser beam colored in red depicts the gradual reduction in laser spot and in turn the effective stem piercing diameter at $7$ different locations, spot diameters as such ranging approximately from $3.8mm$ to $0.1mm$. $10$ experimental iterations are performed per spot diameter making a total of $70$ trials for the coarse tuning set of experiments, videos of which are available for the reader online\footnote{\url{https://www.youtube.com/watch?v=MUS7bZu477s}}. This experiment aims at computing the average piercing velocity $\overline{v}_p$ of each laser spot diameter candidate, where the recorded videos are used to time the duration from engaging the laser power till the moment of laser piercing the stem designating the piercing time $t_p$ in seconds. A calibrated ruler is also used to measure the stem diameter ${\phi}_s$ in millimeters assuming a cylindrical stem, with both information in hand $\overline{v}_p$ is computed. As previously concluded, the average stem piercing velocity $\overline{v}_p$ alone is not indicative of the actual speed of cut, however when multiplied with the spot diameter $\phi_{ls}$ we obtain what we refer to in the sequel by the stem pierce constant $C_p$:
\begin{equation*}
    C_p = \overline{v}_p*\phi_{ls},
\end{equation*}
which gives a more accurate figure of the stem-area etching speed. The results for the coarse tuning set of experiments are provided in Table \ref{table_lateral_spot_diameter_vs_cut_velocity}. In this table we observe that although a laser spot diameter of $0.09mm$ is $44$ times faster than a $3.79mm$ spot in piercing velocity, it can be equally bad in stem etching as indicated by an almost identical value of pierce constant $C_p$. The spot diameter value with most promising result is $0.71mm$ will be used for fine tuning in a following set of identical experiments. 

\renewcommand{\arraystretch}{1.1} 
\begin{table}[b]
\captionsetup{aboveskip=0pt,belowskip=-10pt}
\caption{Results of fine tuning the laser spot diameter $\phi_{ls}$,with best performing in bold. }
\label{table_lateral_spot_diameter_optimization}
\begin{center}
\begin{tabular}{|c|c||c||c|c|}
\hline
$\phi_{ls}(mm)$ & $\overline{\phi}_s(mm)$ & $\overline{t}_p(s)$ & $\overline{v}_p(mm/s)$ & $C_p(mm^2/s)$ \\
\hline
\hline
$0.5$ & $2.1$  & $1.16$  & $1.81$     & $0.90$ \\
$0.6$ & $2.4$  & $1.34$  & $1.79$     & $1.07$ \\
$0.8$ & $2.1$  & $1.50$  & $1.4$     & $1.14$ \\
$\textbf{0.9}$ & $\textbf{2.2}$  & $\textbf{1.47}$  & $\textbf{1.49}$     & $\textbf{1.36}$ \\
$1.0$ & $2.1$  & $1.74$  & $1.21$     & $1.23$ \\
$1.1$ & $2.3$  & $2.52$  & $0.91$     & $1.02$ \\
\hline
\end{tabular}
\end{center}
\end{table}

The final results of the laser spot diameter fine tuning are provided in Table \ref{table_lateral_spot_diameter_optimization} for $6$ more diameter values, videos of those experiments are also available online\footnote{\url{https://www.youtube.com/watch?v=rR-wyuchMrM}}. One initial observation is that the near optimal value $0.9mm$ has a pierce constant less than that of the $0.71mm$ in the previous set of experiments. This is because the stem specimens used are kept in storage at low temperature, which are then prepared for the experiments and left in room temperature for varying durations, that change from one set of experiments to another.

\begin{algorithm}[b!]
\caption{Strawberry localization algorithm}
 \begin{algorithmic}[1]
 \renewcommand{\algorithmicrequire}{\textbf{Input:}}
 \renewcommand{\algorithmicensure}{\textbf{Output:}}
 \REQUIRE colored scene point cloud $^{c_1}\mathcal{C}_{s_1}$ from RGB-D CAM1, \\
          \hspace{4.5mm} and $^{c_2}\mathcal{C}_{s_2}$ from RGB-D CAM2. \\
 \ENSURE  bounding box coordinate vectors for localized strawberries in harvester base frame.
  \STATE $^{b}\mathcal{C}_{s_1} = {^{b}\bm{\textbf{T}}_{c_1}} ^{c_1}\mathcal{C}_{s_1}$
  \STATE$^{b}\mathcal{C}_{rs_1} = \texttt{EXTR(}^{b}\mathcal{C}_{s_1},-0.3, 0.3, -0.2, 0.2, 0.5, 0.7\texttt{)}$
  \STATE$^{b}\mathcal{C}_{cp_1} = \texttt{EXTR(}^{b}\mathcal{C}_{s_1},-0.09,-0.07, 0.1,0.11, 0.3,0.35\texttt{)}$
  
  \STATE $^{b}\mathcal{C}_{red_1} = \emptyset$
  \FORALL{point $^{b}{c}^{i}_{p_1}$ in $^{b}\mathcal{C}_{rs_1}$}
   \IF {($|^{b}{c}^{i}_{rs_1}.r-{^{b}{\overline{c}}^{}_{cp_1}}.r| < r_{th}$ \textbf{and} $|^{b}{c}^{i}_{rs_1}.g-{^{b}{\overline{c}}^{}_{cp_1}}.g| < g_{th}$ \textbf{and} $|^{b}{c}^{i}_{rs_1}.b-{^{b}{\overline{c}}^{}_{cp_1}}.b| < b_{th}$ )}
    \STATE $^{b}\mathcal{C}_{red_1} \gets {^{b}{c}^{i}_{rs_1}}$
   \ENDIF
  \ENDFOR
  
  \STATE $^{b}\mathcal{C}_{red} = {^{b}\mathcal{C}_{red_1}} + {^{b}\mathcal{C}_{red_2}} $
  
  \STATE $^{b}\mathcal{C}_{straw} = \texttt{EuCS(}t, s_{min}, s_{max}, ^{b}\mathcal{C}_{red} \texttt{)}$
  \FORALL{cluster $^{b}\textbf{c}^{i}_{straw}$ in $^{b}\mathcal{C}_{straw}$}
    \STATE $^{b}\textbf{x}_{straw}^{min} \gets \texttt{MIN(} ^{b}\textbf{c}^{i}_{straw}, \texttt{0)}$
    \STATE $^{b}\textbf{x}_{straw}^{max} \gets \texttt{MAX(} ^{b}\textbf{c}^{i}_{straw}, \texttt{0)}$
    \STATE $^{b}\textbf{y}_{straw}^{min} \gets \texttt{MIN(} ^{b}\textbf{c}^{i}_{straw}, \texttt{1)}$
    \STATE $^{b}\textbf{y}_{straw}^{max} \gets \texttt{MAX(} ^{b}\textbf{c}^{i}_{straw}, \texttt{1)}$
    \STATE $^{b}\textbf{z}_{straw}^{min} \gets \texttt{MIN(} ^{b}\textbf{c}^{i}_{straw}, \texttt{2)}$
    \STATE $^{b}\textbf{z}_{straw}^{max} \gets \texttt{MAX(} ^{b}\textbf{c}^{i}_{straw}, \texttt{2)}$
  \ENDFOR
 \RETURN $^{b}\textbf{x}_{straw}^{min}, {^{b}\textbf{x}_{straw}^{max}}, {^{b}\textbf{y}_{straw}^{min}}, {^{b}\textbf{y}_{straw}^{max}},$\\ \hspace{10.8mm} $^{b}\textbf{z}_{straw}^{min}, {^{b}\textbf{z}_{straw}^{max}}$
 \end{algorithmic}
\label{strawberry_localization_algorithm}
\end{algorithm}

\begin{algorithm}[b!]
 \caption{\texttt{EXTR()}}
 \begin{algorithmic}[1]
 \renewcommand{\algorithmicrequire}{\textbf{Input:}}
 \renewcommand{\algorithmicensure}{\textbf{Output:}}
 \REQUIRE point cloud $^{}\mathcal{C}_{}$, \\
          \hspace{4.2mm} spatial limits in order $x^{+}_{l}$, $x^{-}_{l}$, $y^{+}_{l}$, $y^{-}_{l}$, $z^{+}_{l}$, $z^{-}_{l}$. \\
 \ENSURE  a subset point cloud $^{}\mathcal{C}_{sl} \subset ^{}\mathcal{C}_{} $ of all points located within the spatial limits.
  \FORALL{point $^{}{c}^{i}_{}$ in $^{}\mathcal{C}_{}$}
   \IF {($^{}{c}^{i}_{}.x < x^{+}_{l}$ \textbf{and} $^{}{c}^{i}_{}.x > x^{-}_{l}$ \textbf{and} $^{}{c}^{i}_{}.y < y^{+}_{l}$ \textbf{and} $^{}{c}^{i}_{}.y > y^{-}_{l}$ \textbf{and} $^{}{c}^{}_{s}.z < z^{+}_{l}$ \textbf{and} $^{}{c}^{i}_{}.z > z^{-}_{l}$ )}
   \STATE $^{}\mathcal{C}_{sl} \gets {^{}{c}^{i}}_{}$
   \ENDIF
  \ENDFOR
 \RETURN $^{}\mathcal{C}_{sl}$
 \end{algorithmic}
\label{spatial_extraction_algorithm}
\end{algorithm}

\section{Control Algorithm and Harvesting Demo}

\begin{figure*}[t!]
    \centering
    \includegraphics[width=\textwidth]{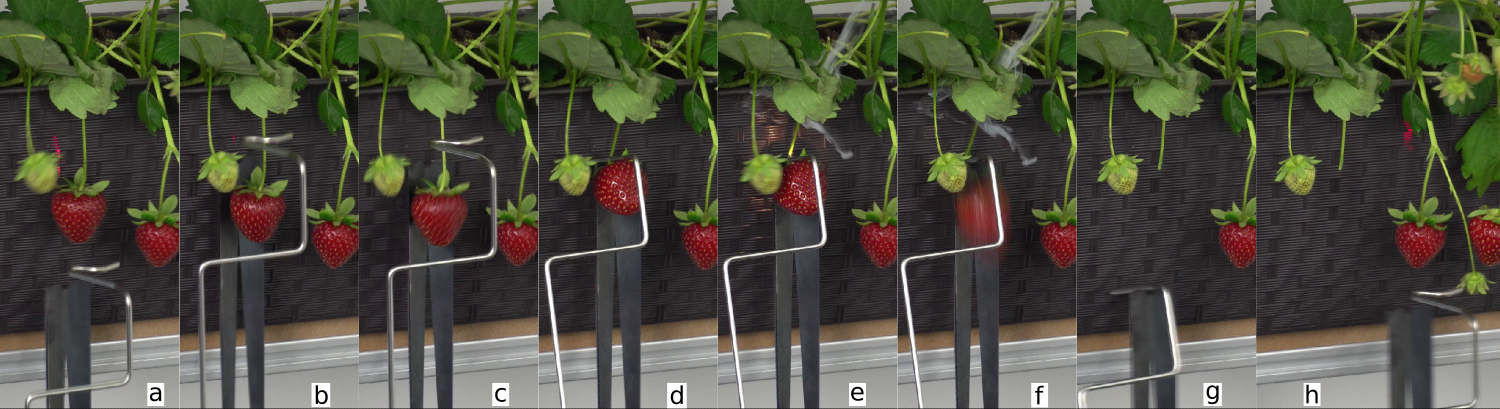}
    \captionsetup{aboveskip=10pt}
    \captionsetup{belowskip=-10pt}
    \caption{Snapshots of the demo experiment autonomously harvesting strawberries featuring the developed tool. The smooth and slim trapper minimize disturbance to the fruit with safe handling as in (d). It also bears the laser heat in (e) to prevent disease transmission.  }
    \label{figure_full_stroke_steps}
\end{figure*}

In this section we present the harvesting experimental demo and details of the pseudo algorithms employed for strawberry segmentation, localization and harvester motion control. In such demo, the system described in the previous section is used to harvest a collection of $11$ strawberries in full autonomy. The point cloud of these shown in Fig. \ref{figure_strawberry_segmentation_point_cloud}. The hardware used feature the $3$ DOF linear motion system, two realsense D415 depth cameras, and a $50$ Watt Raycus RFL-P50QB module \cite{fiber_laser} as the fiber laser source. The harvesting tool is 3D printed except for the stem-trapper and the trapping-groove that are subjected to extensive laser heat. Of the shelf microcontroller development board is used to control the convex lens and trapper movements, the laser activation, as well as monitoring the strawberry detachment photo interrupters. On the software side, we use the realsense SDK library for interfacing with the cameras, and the Point Cloud Library (PCL) \cite{Rusu_ICRA2011_PCL} to process the resulting RGB-D points. A video of the harvesting experiment is submitted with this paper and available online\footnote{ \url{https://www.youtube.com/watch?v=Z7qEVdkhq2s}.}, snapshots of which are shown in Fig. \ref{figure_full_stroke_steps}.

\subsection{Strawberry Localization}
\begin{figure}[t!]
    \centering
    \includegraphics[width=1.0\linewidth]{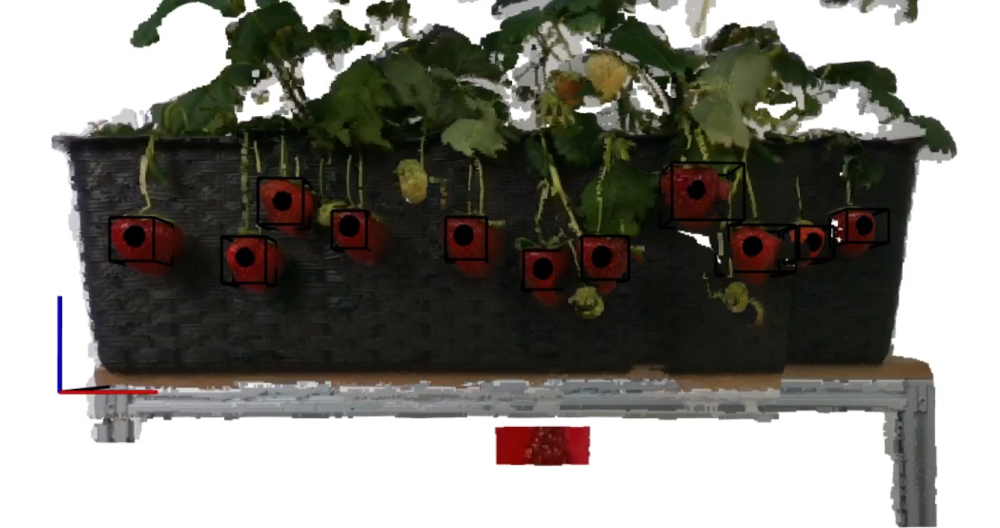}
    \captionsetup{aboveskip=10pt}
    \captionsetup{belowskip=-10pt}
    \caption{Reduced, dual RGB-D camera augmented scene point cloud showing strawberry segmentation bounding boxes together with the calibration pattern. Resulting from pseudo-code in Algorithm \ref{strawberry_localization_algorithm}.}
    \label{figure_strawberry_segmentation_point_cloud}
\end{figure}
Although fruit segmentation is out of scope of this work, it is essential for a fully autonomous demonstration of the system. Therefore, we present a simple approach for the limited purpose of demo completeness. The authors acknowledge that numerous prior work based on deep learning methods \cite{Ge_2019,Zhichao_2025} would achieve better results in terms of strawberries growing in clusters. The pseudo-code for strawberry localization is provided in Algorithm \ref{strawberry_localization_algorithm}. It takes as input the raw scene point cloud acquired by both RGB-D cameras $^{c_1}\mathcal{C}_{s_1}$, and $^{c_2}\mathcal{C}_{s_2}$ expressed in the corresponding camera frames. These are then transformed to the harvester base frame (code line 1) to form the scene point clouds $^{b}\mathcal{C}_{s_1}$ and $^{b}\mathcal{C}_{s_2}$ with the latter omitted from Algorithm \ref{strawberry_localization_algorithm} and in the sequel to avoid redundancy. From which, a \textit{reduced scene} cloud set $^{b}\mathcal{C}_{rs_1} \subset {^{b}\mathcal{C}}_{s_1}$ is then constructed, where a point ${^{b}{c}^{i}}_{s_1}$ in the scene cloud $^{b}\mathcal{C}_{s_1}$ is added to the reduced cloud if it resides in a spatial window characterized by the maximum and minimum limits $x^{+}_{l}, x^{-}_{l}, y^{+}_{l}, y^{-}_{l}, z^{+}_{l}, z^{-}_{l}$ in $x, y,$ and $z$ coordinates respectively of the harvest tool base frame, and computed using the spatial extraction $\texttt{EXTR()}$ (code line 2) in Algorithm \ref{spatial_extraction_algorithm}. This reduces the forthcoming computation to the area of interest that lies within the workspace of the linear harvesting tool, such reduced point cloud is shown in Fig. \ref{figure_strawberry_segmentation_point_cloud}.
Spacial extraction $\texttt{EXTR()}$ is again used (code line 3) to extract the \textit{calibration pattern} point cloud $^{b}\mathcal{C}_{cp_1}$ or real fruit color from a specified location as per the harvest tool design in Fig. \ref{figure_harvesting_tool_CAD}. Ripe strawberries are simply extracted by thresholding the red color using the RGB variation thresholds $r_{th}, g_{th}, b_{th}$ against those of the weighted average of the calibration pattern point cloud $^{b}\mathcal{\overline{C}}_{cp_1}$ to form the \textit{red} point cloud $^{b}\mathcal{C}_{red_1} \subset {^{b}\mathcal{C}}_{rs_1}$ (code lines 4 to 9). Calibration and red color extraction had to be done separately per RGB-D camera, the resulting red points can now be augmented in $^{b}\mathcal{C}_{red}$ (code line 10). We then use Euclidean cluster segmentation algorithm to segment each individual strawberry. The output of the function $\texttt{EuCS(}t, s_{min}, s_{max}, ^{b}\mathcal{C}_{red} \texttt{)}$ is a set of strawberry clusters (set of point clouds each representing a single strawberry) $^{b}\mathcal{C}_{straw}$ arranged in ascending order of the y-coordinate values. With $t, s_{min}, s_{max}$ denoting the segmentation tolerance, minimum and maximum cluster sizes respectively.  The $\texttt{MIN(} ^{b}\textbf{c}^{i}_{straw}, \texttt{0)}$ function, supplied with a point cloud $^{b}\textbf{c}^{i}_{straw}$ and an index, will return the minimum value available at such index in all point cloud points, whereas, supplied with a vector, will return the smallest value irrespective of the index. The lengthy pseudo-code for the aforementioned two functions is omitted for convenience. The output is a set of vectors defining the bounding boxes of the localized strawberries, these are depicted in Fig. \ref{figure_strawberry_segmentation_point_cloud}.

\subsection{Harvesting Demo}
Given a list of fruit bounding boxes, the motion controller computes the geometric center of each strawberry, and given the depth of each in the list, motion commands are generated for the harvesting tool to move to the $x$, $y$ centroid coordinates of each fruit at a $30mm$ lower elevation in the $z$ axis below the lower bound of the corresponding bounding box as shown in Fig. \ref{figure_full_stroke_steps}(a). The moving frame in which these motion commands are defined lies in the middle of the trapping groove (the reader is referred to Fig. \ref{figure_harvesting_tool_CAD}). The tool then rises in $z$ axis to surpass the top elevation of the respective bounding box by $20mm$ as in Fig. \ref{figure_full_stroke_steps}(b), then retracts towards the fruit which then comes in stable contact with the v-shaped trapping groove as depicted in Fig. \ref{figure_full_stroke_steps}(c). The polished chrome stainless steel trapper then moves to trap the stem smoothly with minimal friction as in Fig. \ref{figure_full_stroke_steps}(d), improving on the previous work by the authors \cite{Sorour2023}. The fiber laser is then energized, as well as the lens lateral motion system to etch through the now trapper stem in Fig. \ref{figure_full_stroke_steps}(e). The stem is eventually cut and the fruit is detached as in Fig. \ref{figure_full_stroke_steps}(f), but the laser will not deactivate until the free falling fruit is detected by one of the triple infrared photo interrupters illustrated in \ref{figure_harvesting_tool_CAD}(d). When the fruit is detected and the laser power is deactivated, the tool descends in $z$ axis until it is $30mm$ below the lower side of the next-in-queue bounding box before moving to the $x$, $y$ coordinate value of its centroid as shown in Fig. \ref{figure_full_stroke_steps}(g), Fig. \ref{figure_full_stroke_steps}(h) respectively. This marks the end of a single cutting cycle. The localization algorithm consumes $100ms$ in worst case scenario on a standard laptop with intel core-i$7$ ${8^{th}}$ generation processor.

The average cycle and stem-cut time is $5.56$, and $2.88$ seconds respectively, with the former being the addition of the linear movement and the laser cutting durations. Due to the fact that strawberries used in the demo are stored at low temperature, the aforementioned cutting time is expected to be shorter in real world application. In the supplied demo video, the authors provide the output of a thermal camera to show the heat applied using laser cutting. Although the sensor accuracy and frame rate are low, it is evident that temperatures at interaction can reach up to $188$ degrees Celsius and that the tool is always hot, which highlights a unique advantage of the presented approach, namely preventing the transmission of local plant diseases. Robustness against light intensity/source variation is presented in another video experiment here\footnote{ \url{https://www.youtube.com/watch?v=rjt0wkRDObg}.} thanks to the adoption of active color calibration. In this experiment strawberries are successfully segmented/localized in artificial low and medium indoor lighting as well as indirect sun light source. This is highly relevant since table-top grown produce harvesting is done in both lighting conditions at night and day times respectively. Although deep learning approaches provide promising results to this problem, active color calibration is expected to greatly enhance the results, especially with green hanging produce such as peppers. Exploring such methods to tackle strawberries growing in clusters motivates future work as a final step before real farm deployment of our system.

\section{Conclusion}
In this work, a customized harvesting prototype has been presented for table-top grown strawberries. It employs a low friction trapping tool to force the stem into a pre-defined location where a focused laser beam performs the cut. The v-shaped trapper is robust to fruit localization errors of up to $20$mm. Extensive experiments are performed to optimize the laser beam features against the cutting time. Temperatures of up to $188\degree$ Celsius are reported in the interaction between the tool and the plant, effectively preventing the spread of local diseases. Successful demonstration of indoor strawberry harvesting validated the proposed approach.


\bibliographystyle{IEEEtran}  
\bibliography{main}

\end{document}